\documentclass[conference]{IEEEtran}
\IEEEoverridecommandlockouts

\usepackage{subcaption}
\usepackage{cite}
\usepackage{amsmath,amssymb,amsfonts}
\usepackage{algorithmic}
\usepackage{graphicx}
\usepackage{textcomp}
\usepackage{xcolor}
\usepackage{mathrsfs}
\usepackage{booktabs}
\usepackage{url}
\usepackage{eso-pic}
\usepackage{booktabs}
\usepackage{tabularx}
\usepackage{array}
\usepackage{xcolor}

\usepackage{xcolor}
\usepackage{booktabs}
\usepackage{tabularx}
\usepackage{array}

\newcolumntype{Y}{>{\raggedright\arraybackslash}X}

\newcommand{\casehit}[1]{\textcolor{blue!70!black}{#1}}
\newcommand{\casefp}[1]{\textcolor{red!75!black}{#1}}
\newcommand{\casemiss}[1]{\textcolor{orange!90!black}{#1}}

\newcommand{\predcell}[3]{%
\begin{tabular}[t]{@{}p{0.13\linewidth}@{}p{0.84\linewidth}@{}}
\textbf{#1} & #2 \\
& {\scriptsize\textit{Miss:} #3}
\end{tabular}
}

\def\BibTeX{{\rm B\kern-.05em{\sc i\kern-.025em b}\kern-.08em
    T\kern-.1667em\lower.7ex\hbox{E}\kern-.125emX}}
\begin{document}

\title{HEM-Med: A Hierarchical Agentic Memory Framework for Safe Multi-Center Medication Recommendation}

\author{
\IEEEauthorblockN{
Min Xiao\textsuperscript{1,2,\textdagger},
Jin Li\textsuperscript{4,\textdagger},
Shoujin Wang\textsuperscript{4},
Yeguang Zhen\textsuperscript{1,2},
Yishuo Li\textsuperscript{1,2},
Wenpeng Lu\textsuperscript{1,2,3,\textcolor{red}{*}}
}

\IEEEauthorblockA{
\textsuperscript{1}
Key Laboratory of Computing Power Network and Information Security, Ministry of Education;\\
Shandong Computer Science Center (National Supercomputer Center in Jinan);\\
Qilu University of Technology (Shandong Academy of Sciences);\\
\textsuperscript{2}Shandong Provincial Key Laboratory of Computing Power Internet and Service Computing;\\
Shandong Fundamental Research Center for Computer Science, Jinan, Shandong, China\\
\textsuperscript{3}Shandong Academy of Artificial Intelligence, Jinan, Shandong, China\\
Email: \texttt{min.xiao02@foxmail.com}; \texttt{zhenyeguang92@gmail.com};\\
\texttt{yishuoli@foxmail.com}; \texttt{wenpeng.lu@qlu.edu.cn}
}

\IEEEauthorblockA{
\textsuperscript{4}
University of Technology Sydney\\
Email: \texttt{Jin.Li-4@student.uts.edu.au}; \texttt{shoujinwang@foxmail.com}
}

\IEEEauthorblockA{
\textsuperscript{\textdagger}
 Equal Contributors
\textsuperscript{\textcolor{red}{*}}
Corresponding author
}
}

\maketitle

\begin{abstract}
Medication recommendation (MR) is critical for clinical decision support. While existing MR methods are mostly designed for single-center learning, where only one independent hospital is involved, multi-center MR with hospitals of different scales has become increasingly demanding in practice. In multi-center settings, abundant data from large medical centers can improve recommendation performance for low-resource centers. Despite a few preliminary attempts on this task, two critical gaps remain. Gap 1: Existing multi-center methods fail to disentangle transferable cross-hospital clinical experience from hospital-specific prescribing patterns, limiting clinical feasibility and physician acceptance. Gap 2: They commonly overlook multi-label safety constraints of MR, including drug-drug interaction risks and local formulary feasibility, which are essential for safe real-world deployment. To address these gaps, we propose HEM-Med, a novel hierarchical agentic memory framework for multi-center MR. To address Gap 1, we design a disentangled hierarchical memory module that builds patient-level medication-context attribution records and disentangles them into cross-hospital experience memory and hospital-specific residual memory. 
To address Gap 2, we introduce a safety-aware medication set generation module, which integrates dual-level experience retrieval with safety-aware set decoding to produce clinically relevant, safer, and locally adaptive medication recommendations. Experiments on a real-world multi-center eICU dataset show that HEM-Med outperforms representative transfer learning, LLM-based, and MR baselines in terms of both recommendation accuracy and safety.
\footnote{Our code is publicly available at: \url{https://github.com/xiaointo/HEM-Med}.
}

\end{abstract}


\section{Introduction}
Medication recommendation (MR) is a key task in clinical decision support systems. The goal is to analyze a patient's complex Electronic Health Record (EHR) and generate a set of candidate medications that are well aligned with the patient’s current physiological status and therapeutic needs~\cite{shang2019gamenet,yang2021safedrug}. 

Early MR studies focused on model development within a single center, where training and evaluation are conducted using EHR data from the same hospital. Representative methods include GAMENet, SafeDrug, and DKGM~\cite{shang2019gamenet,yang2021safedrug,zhao2025dkgm}. Despite their effectiveness in single-center settings, these methods rely on local data and may perform poorly in low-resource hospitals owing to data sparsity and cold-start issues~\cite{Rumor_zhangqi,wang2026eval4}.

\begin{figure}[t]
\centering
\includegraphics[width=0.9\linewidth]{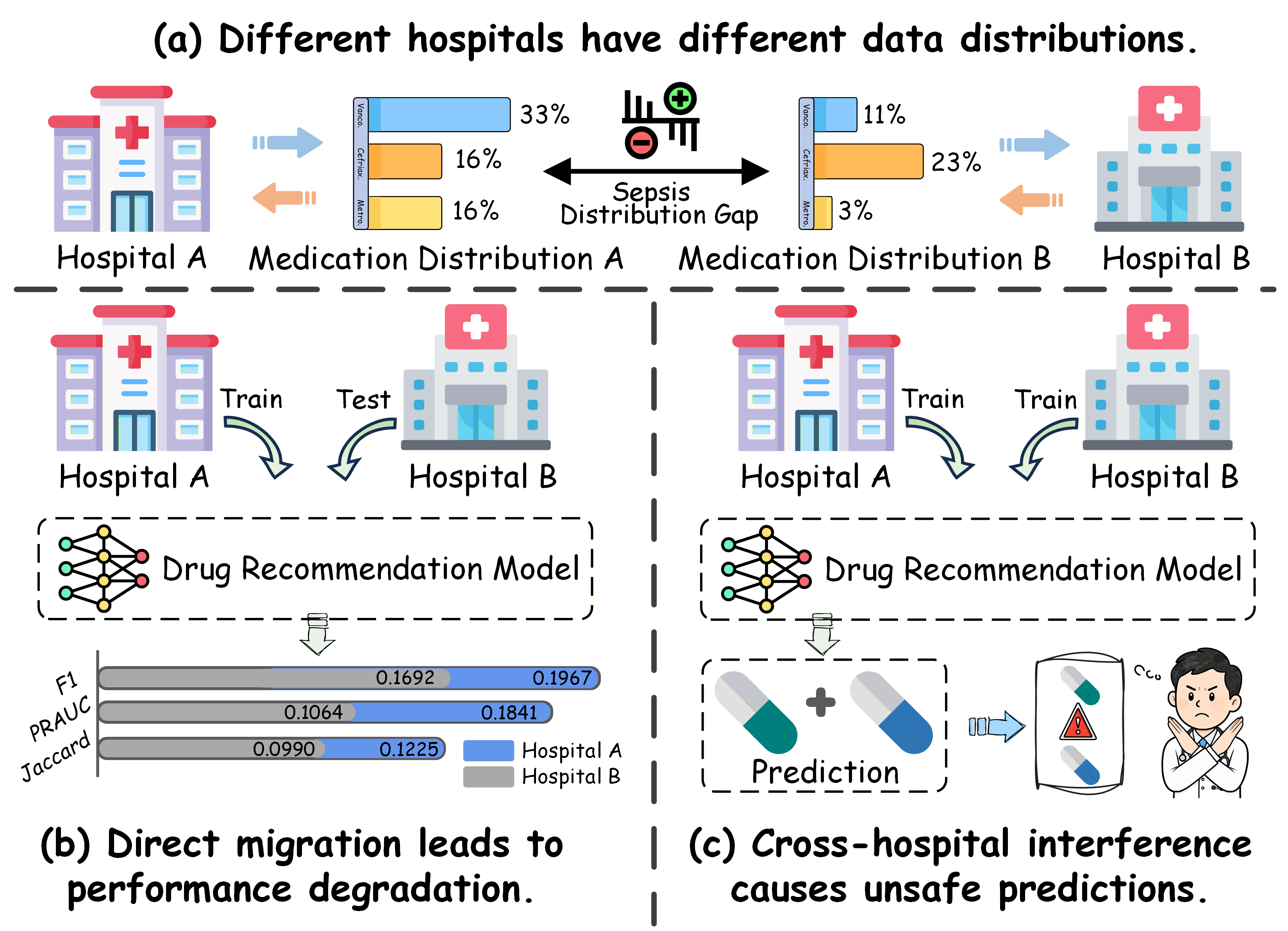}
\caption{Motivating example of multi-center medication recommendation. Heterogeneous hospital prescribing patterns may cause negative transfer, and ignoring local feasibility and drug safety may lead to unsafe medication combinations.} \label{fig:gap}
\vspace{-3mm}
\end{figure}
Recent studies address single-center limitations through two lines of research: cross-domain and multi-center transfer for cross-hospital knowledge sharing. The first line borrows ideas from General Cross-Domain Recommendation (CDR), which transfers or shares knowledge across related domains, such as EMCDR and DDTCDR~\cite{man2017cross,li2020ddtcdr}. The second line targets multi-center healthcare scenarios. A representative method TEMPT~\cite{liu2025contrastive} pre-trains on pooled multi-center prescription data and then relies solely on prompt fine-tuning to adapt to the target hospital distribution~\cite{wang2020modelling,WSD_Fanqing}.


Despite recent progress, existing studies still suffer from two major gaps in multi-center medication recommendation.
\textbf{Gap 1: Entangled cross-hospital knowledge.}
Existing multi-center MR methods often overlook that prescribing decisions reflect both generalizable medical knowledge and hospital-specific resources, clinical protocols, and prescribing preferences. As shown in Fig.~\ref{fig:gap} (a), medication distributions may vary across hospitals for the same disease~\cite{liu2025contrastive}. Fig.~\ref{fig:gap} (b) further illustrates that transferring entangled cross-hospital knowledge in ICU settings with coexisting conditions may cause negative transfer and impair recommendation performance.
\textbf{Gap 2: Insufficient safety awareness.}
Existing transfer-based methods prioritize representation alignment but often overlook clinical safety constraints. Unlike general CDR, MR requires multi-label predictions that satisfy treatment needs, DDI constraints, and local formulary availability~\cite{yang2021safedrug,tatonetti2012data,wishart2018drugbank}. As illustrated in Fig.~\ref{fig:gap} (c), ignoring these constraints may yield inappropriate or unsafe medication combinations.

To address these gaps, we propose HEM-Med, a hierarchical agentic memory framework for multi-center MR. For Gap 1, a disentangled memory module separates transferable clinical experience from hospital-specific prescribing patterns into cross-hospital experience memory and hospital-specific residual memory. For Gap 2, a safety-aware medication set generation module combines dual-level retrieval with DDI-aware set decoding to generate clinically relevant, safer, and locally adaptive recommendations.

The contributions are summarized as follows:
\begin{itemize}
    \item We propose HEM-Med, a hierarchical agentic memory framework for multi-center MR that explicitly disentangles cross-hospital experience from hospital-specific prescribing preferences.

    \item We develop a safety-aware module for ICU medication set generation, integrating dual-level retrieval and DDI-aware decoding to account for clinical relevance, local feasibility, and DDI risks.

    \item Experiments on the eICU dataset show that HEM-Med improves recommendation accuracy while maintaining competitive medication safety.
\end{itemize}

\section{Related Work}

\subsection{Medication Recommendation}\label{AA}

EHR-based MR aims to recommend personalized, clinically safe medication combinations. Existing approaches include longitudinal modeling with RETAIN and DMNC~\cite{choi2016retain, le2018dual}, time-aware modeling with MR-DTR~\cite{li2025time}, and DDI-aware recommendation with GAMENet, SafeDrug, and MoleRec~\cite{shang2019gamenet,yang2021safedrug,yang2023molerec}. DKINet, TCMPR, and SECGI~\cite{liu2024dkinet,dong2022tcmpr,yang2025secgi} enhance patient representations with external knowledge or semantics. However, most methods target single-center settings and may suffer from negative transfer across hospitals~\cite{lu2020concept,zhou2025large}.

\subsection{Cross-Domain Transfer and Multi-Center Learning}
CDR and multi-center transfer alleviate data sparsity through knowledge sharing. EMCDR~\cite{man2017cross} and DisenCDR~\cite{cao2022disencdr} align domain representations, while TEMPT~\cite{liu2025contrastive} adapts to hospitals through pre-training and prompt-tuning. However, general CDR overlooks combination therapy and DDI risks, while medical transfer lacks explicit consensus–preference disentanglement, risking negative transfer~\cite{zhangsecon}.

\subsection{LLM Agents and Knowledge-Augmented Healthcare}
LLMs support clinical decision-making through semantic reasoning~\cite{singhal2023large}. Knowledge-enhanced methods include multi-agent collaboration (MedAgents, MDTeamGPT~\cite{tang2024medagents,chen2025mdteamgpt}) and retrieval- or tool-augmented reasoning (ClinicalGPT, TxAgent, JointCoder, SOAPTriage~\cite{wang2023clinicalgpt,gao2025txagent,liu2026joint,wang2026soaptriage}). However, their prompt-based knowledge integration often lacks explicit separation of general medical safety principles from hospital-specific execution rules for multi-center MR.

\section{Preliminaries}

\subsection{Multi-Center Electronic Health Records Data}

Multi-center EHR data comprise independent ICU records from multiple hospitals. Let $\mathcal{H}=\{h_1,\ldots,h_{|\mathcal{H}|}\}$ denote the hospital set, and $\mathcal{D}$, $\mathcal{P}$, $\mathcal{O}$, and $\mathcal{M}$ the diagnosis, procedure, abnormal laboratory signal, and standardized medication vocabularies, respectively. The $i$-th record is represented as:
\begin{equation}
    \mathbf{r}_i=(h_i,D_i,P_i,O_i,M_i),
\end{equation}
where $h_i\in\mathcal{H}$ is the hospital identifier, $D_i\subseteq\mathcal{D}$, $P_i\subseteq\mathcal{P}$, $O_i\subseteq\mathcal{O}$, and $M_i\subseteq\mathcal{M}$ denote the observed diagnoses, procedures, abnormal laboratory signals, and medications, respectively.

\subsection{Clinical Anchors}

The clinical anchor set for EHR instance $i$ comprises observed diagnoses, procedures, and abnormal laboratory signals:
\begin{equation}
\begin{aligned}
    A_i =
    &
    \{(\mathrm{diag}, d) \mid d \in D_i\}
    \cup
    \{(\mathrm{proc}, p) \mid p \in P_i\}\\
    &\cup
    \{(\mathrm{lab}, o) \mid o \in O_i\}.
\end{aligned}
\end{equation}
Each anchor $a \in A_i$ represents a typed clinical signal: a diagnosis, procedure, or abnormal laboratory state.

\subsection{Drug-Drug Interaction Knowledge}

We denote the drug-drug interaction knowledge as a graph:
\begin{equation}
    \Omega_{\mathrm{DDI}} = (\mathcal{M}, \mathcal{E}_{\mathrm{DDI}}),
\end{equation}
where $\mathcal{M}$ is the standardized medication vocabulary and $\mathcal{E}_{\mathrm{DDI}} \subseteq \mathcal{M} \times \mathcal{M}$ contains potentially interacting medication pairs. Each node represents a medication, and each edge $(u,v)$ represents a potential interaction with severity score $\mathrm{sev}(u,v)$.

\subsection{Medication Recommendation}

Based on the observed clinical context of the $i$-th EHR instance, we define the model input as:
\begin{equation}
    \mathbf{x}_i=(h_i,D_i,P_i,O_i),
\end{equation}
while $M_i$ is the ground-truth set. Multi-center MR estimates conditional relevance scores $s(m\mid\mathbf{x}_i)$ for $m\in\mathcal{M}$ and decodes the medication set $\hat{Y}_i\subseteq\mathcal{M}$ as:
\begin{equation}
    \hat{Y}_i=
    \operatorname{Decode}
    \left(
    \{s(m\mid\mathbf{x}_i)\}_{m\in\mathcal{M}},
    \Omega_{\mathrm{DDI}},
    \mathbf{x}_i
    \right).
\end{equation}

\section{Method}
\begin{figure*}[t]
    \centering
    \includegraphics[width=0.8\textwidth]{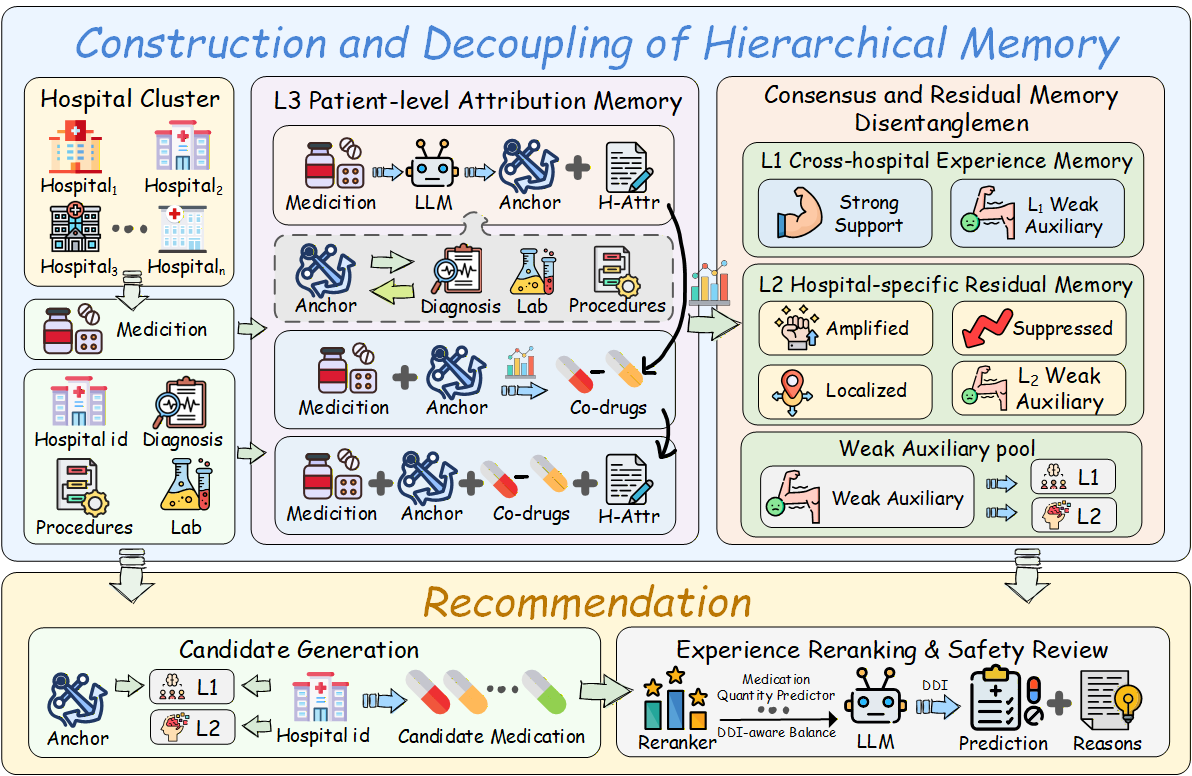}
    \caption{Overall agentic framework of HEM-Med. HEM-Med retrieves disentangled cross-hospital and hospital-specific experience, and generates medication sets through calibrated prediction and DDI-aware decoding.}
    \label{fig:evodrug_mas_framework}
    \vspace{-3mm}
\end{figure*}

We propose HEM-Med, a hierarchical agentic memory framework for multi-center MR. As shown in Fig.~\ref{fig:evodrug_mas_framework}, it comprises two components: (1) consensus–residual memory disentanglement, which separates historical medication–context experience into cross-hospital consensus and hospital-specific residual memories; and (2) safety-aware recommendation, which combines dual-level retrieval, experience-aware reranking, DDI-aware set decoding, and safety review to generate locally feasible and clinically safe medication sets.

\subsection{Patient-Level Attribution Memory}

Raw ICU EHRs lack explicit medication links to diagnoses, procedures, or laboratory abnormalities. HEM-Med builds auditable $L_3$ patient-level memory via constrained DeepSeek-assisted attribution.

For each training ICU stay $s_i$ and each observed medication $m_j \in M_i$, we construct a stay-medication query:
\begin{equation}
q_{ij} =
\left(
h_i, m_j, D_i, P_i, O_i
\right),
\end{equation}
where $h_i$ denotes the hospital identifier,  and $D_i$, $P_i$, and $O_i$ denote the observed diagnoses, procedures, and abnormal laboratory signals of the current stay, respectively.

DeepSeek-v4-flash selects at most one observed context anchor for constrained attribution, without adding external indications or clinical facts:
\begin{equation}
\begin{aligned}
a_{ij} \in
&
\{\mathrm{diag}(d) \mid d \in D_i\}
\cup
\{\mathrm{proc}(p) \mid p \in P_i\}\\
&\cup
\{\mathrm{lab}(o) \mid o \in O_i\}
\cup
\{\emptyset\}.
\end{aligned}
\end{equation}
Insufficient context yields $\emptyset$. Attribution is a conservative empirical hypothesis, not a verified prescribing intention, indication, or causal treatment relation.

For each validated stay-medication pair, the $L_3$ record is defined as:
\begin{equation}
e^{(L_3)}_{ij}
=
\left(
s_i, h_i, m_j, a_{ij}, r_{ij}, S_{ij}
\right),
\end{equation}
where $r_{ij}$ denotes the conservative attribution hypothesis, and $S_{ij}$ stores training-set statistical support for subsequent aggregation.

Training-only support statistics distinguish cross-hospital experience from hospital-specific residual patterns in anchor–medication associations.

All validated $L_3$ training records are collected as:
\begin{equation}
\mathcal{K}^{(L_3)}
=
\left\{
e^{(L_3)}_{ij}
\mid
s_i \in \mathcal{S}_{\mathrm{train}},
m_j \in M_i
\right\}.
\end{equation}
Three clinicians reviewed 300 randomly sampled $L_3$ records, yielding Fleiss' $\kappa$ values of 0.813 for anchor selection and 0.726 for attribution assessment.

\subsection{Consensus and Residual Memory Disentanglement}


HEM-Med separates patient-level records into cross-hospital experience and hospital-specific residual memories.

\textbf{Cross-hospital experience memory.}
Let $n(a)$ count occurrences of clinical anchor $a$, $n(a,m)$ count its co-occurrences with medication $m$, and $H(a,m)$ count hospitals observing $(a,m)$. The anchor-conditioned medication frequency is:
\begin{equation}
    p(m\mid a)=\frac{n(a,m)}{n(a)}.
\end{equation}
The cross-hospital consensus memory $\mathcal{K}^{(L_1)}$ stores associations that satisfy sufficient support, frequency, and hospital-coverage constraints:
\begin{equation}
\mathcal{K}^{(L_1)}
=
\left\{(a,m)\mid
\mathbf{c}(a,m)\succeq\boldsymbol{\tau},
\ p(m\mid a)\geq\tau_p
\right\}.
\end{equation}
Here, $\mathbf{c}(a,m)=[n(a),n(a,m),H(a),H(a,m)]$
and $\boldsymbol{\tau}=[\tau_a,\tau_{am},\tau_{ha},\tau_h]$.
$H(a)$ denotes the hospital coverage of anchor $a$, and
$\succeq$ denotes element-wise comparison.

\textbf{Hospital-specific residual memory.}
For hospital $h$, anchor $a$, and medication $m$, we define:
\begin{equation}
p_h(m \mid a) = \frac{n_h(a,m)}{n_h(a)},
\end{equation}
where $n_h(a)$ and $n_h(a,m)$ denote the support and
co-occurrence counts within hospital $h$, respectively. The residual
deviation is computed as:
\begin{equation}
\Delta_h(a,m) = p_h(m \mid a) - p(m \mid a).
\end{equation}
Let $\tau^{(L_2)}_a$, $\tau^{(L_2)}_{am}$,
$\tau^{(L_2)}_p$, and $\tau^{(L_2)}_{\Delta}$ be predefined thresholds.  Only local associations with
sufficient hospital-level support and residual magnitude are retained
in $L_2$ memory:
\begin{equation}
\begin{aligned}
\mathcal{K}^{(L_2)} =
\Big\{
&(h,a,m,\Delta_h(a,m),c^{(L_2)}_{h,a,m})
\;\Big|\;
n_h(a) \geq \tau^{(L_2)}_{a},\;\\
&
n_h(a,m) \geq \tau^{(L_2)}_{am}, 
p_h(m \mid a) \geq \tau^{(L_2)}_{p},\;\\
|&\Delta_h(a,m)| \geq \tau^{(L_2)}_{\Delta}
\Big\}.
\end{aligned}
\end{equation}
Here, $c^{(L_2)}_{h,a,m}$ denotes the residual type: amplified, suppressed, or localized. $L_2$ complements $L_1$ by retaining hospital-specific residual prescribing patterns.

\textbf{Weak $L_1$ and Weak $L_2$ Experience.}
HEM-Med aggregates and down-weights medication–context associations below strong $L_1$ or $L_2$ thresholds as weak experience to support inference recall for long-tail medications and rare clinical contexts.

Associations with cross-hospital support below strict $L_1$ thresholds are retained as weak $L_1$ experience:
\begin{equation}
K^{(L_1)}_{\mathrm{weak}}
=
\left\{
(a,m)
\mid
(a,m)\notin \mathcal{K}^{(L_1)},\ 
\mathcal{C}^{(L_1)}_{\mathrm{weak}}(a,m)=1
\right\},
\end{equation}
where \(\mathcal{C}^{(L_1)}_{\mathrm{weak}}(a,m)\) denotes relaxed cross-hospital criteria based on counts, conditional frequency, and hospital coverage.

Similarly, hospital-specific associations with sufficient local support but failing strict $L_2$ residual criteria are retained as weak $L_2$ experience:
\begin{equation}
\begin{aligned}
K^{(L_2)}_{\mathrm{weak}}
=&
\big\{
(h,a,m,\Delta_h(a,m),\rho^{(L_2)}_{h,a,m})
\mid
(h,a,m)\notin \mathcal{K}^{(L_2)},\\
&\mathcal{C}^{(L_2)}_{\mathrm{weak}}(h,a,m)=1
\big\},
\end{aligned}
\end{equation}
where \(\mathcal{C}^{(L_2)}_{\mathrm{weak}}(h,a,m)\) defines relaxed local criteria for support, conditional frequency, and residual deviation.

During retrieval, weak experience is down-weighted and cannot dominate strong $L_1$ or $L_2$ experience:
\begin{equation}
g_{L_1}(m,\mathcal{A}_i)
=
g^{\mathrm{strong}}_{L_1}(m,\mathcal{A}_i)
+
\gamma^{\mathrm{weak}}_{L_1}
g^{\mathrm{weak}}_{L_1}(m,\mathcal{A}_i),
\end{equation}
\begin{equation}
g_{L_2}(m,\mathcal{A}_i,h_i)
=
g^{\mathrm{strong}}_{L_2}(m,\mathcal{A}_i,h_i)
+
\gamma^{\mathrm{weak}}_{L_2}
g^{\mathrm{weak}}_{L_2}(m,\mathcal{A}_i,h_i),
\end{equation}
where $g^{\mathrm{strong}}{L_k}$ and $g^{\mathrm{weak}}{L_k}$ denote strong and weak experience scores, respectively, and $\gamma^{\mathrm{weak}}_{L_k}$ down-weights weak experience, for $k\in{1,2}$.

\subsection{Candidate Generation}
\label{sec:candidate_generation}

Given $x_i=(h_i,D_i,P_i,O_i)$, HEM-Med uses $\mathcal{A}_i=D_i\cup P_i\cup O_i$ to retrieve medication–context experience from $L_1$ cross-hospital and $L_2$ hospital-specific residual memories, forming a candidate medication pool.

For each medication $m$, we compute the memory-based experience score as:
\begin{equation}
s_{\mathrm{mem}}(m \mid x_i)
=
\alpha_1 g_{L_1}(m,\mathcal{A}_i)
+
\alpha_2 g_{L_2}(m,\mathcal{A}_i,h_i),
\end{equation}
where $g_{L_1}$ and $g_{L_2}$ denote the aggregated experience scores from $L_1$ and $L_2$ memory, respectively. 

To reduce the bias toward globally frequent medications, HEM-Med further incorporates prior statistics:
\begin{equation}
\begin{aligned}
\Phi_{\mathrm{prior}}(m,h_i,\mathcal{A}_i)
=&\ 
\phi_g \operatorname{Pr}(m)
+\phi_h \operatorname{Pr}_{h_i}(m) \\
&+\phi_a \operatorname{Pr}(m \mid \mathcal{A}_i)
+\phi_{ha} \operatorname{Pr}_{h_i}(m \mid \mathcal{A}_i),
\end{aligned}
\end{equation}
where $\Phi_{\mathrm{prior}}(m,h_i,\mathcal{A}_i)$ scores medication $m$ using global, hospital, anchor, and hospital–anchor frequencies.

The initial candidate score is defined as:

\begin{equation}
\begin{aligned}
s_{\mathrm{init}}(m \mid x_i)
=&\ 
s_{\mathrm{mem}}(m \mid x_i)
+\beta \Phi_{\mathrm{prior}}(m,h_i,\mathcal{A}_i) \\
&+ b_{\mathrm{comp}}(m)
-\lambda_{\mathrm{cand}}\psi_{\mathrm{cand}}(m),
\end{aligned}
\end{equation}
where $b_{\mathrm{comp}}(m)$ compensates for under-retrieval, and
$\psi_{\mathrm{cand}}(m)$ denotes a DDI-based soft safety penalty.

Top-ranked medications form the candidate pool:
\begin{equation}
B_i
=
\mathrm{Top}_B
\left(
\{s_{\mathrm{init}}(m \mid x_i) \mid m \in \mathcal{M}\}
\right),
\end{equation}
where $B$ is the retained candidate size. The candidate pool $B_i$ is then passed to the learned experience reranker for further calibration and selection.

\subsection{Experience Reranking and Safety Review}

HEM-Med estimates $p_{\mathrm{final}}(m\mid x_i)$ for $m\in B_i$ from calibrated reranking and predicts $p_{\mathrm{size}}(k\mid x_i)$. Safety-constrained DDI-aware set decoding mitigates unsafe or redundant combinations from independent Top-$K$ selection. Here, $\rho(u,v)$ and $\mathrm{sev}(u,v)$ denote empirical co-administration confidence and DDI severity, respectively; $\mathrm{Red}(Y)$ counts redundant medications within therapeutic categories.

HEM-Med generates the final medication set by optimizing the following set-level objective:
\begin{equation}
\begin{aligned}
S(Y \mid x_i)
=
&\sum_{m\in Y}
\log p_{\mathrm{final}}(m \mid x_i)
+
\log p_{\mathrm{size}}(|Y| \mid x_i) \\
&-
\lambda_{\mathrm{DDI}}
\sum_{\substack{u,v\in Y\\u<v}}
\mathrm{sev}(u,v)(1-\rho(u,v))\\
&
-
\lambda_{\mathrm{red}}\mathrm{Red}(Y).
\end{aligned}
\end{equation}

The recommended medication set is obtained as:
\begin{equation}
\hat{Y}_i
=
\arg\max_{Y\subseteq B_i}
S(Y \mid x_i).
\end{equation}

The objective rewards clinical relevance and appropriate set size, penalizes DDI pairs more strongly with greater severity and lower empirical co-administration confidence, and discourages within-category redundancy.

\subsection{Model Training}

HEM-Med separately trains pointwise, pairwise, therapeutic-class, and medication-count submodules for candidate calibration and set-size estimation, combining them only at inference. Given candidates $B_i$ and ground-truth medications $M_i$, each $m\in B_i$ receives a binary label:
\begin{equation}
y_{i,m}=\mathbb{I}(m\in M_i).
\end{equation}
We optimize medication-level prediction with a weighted binary cross-entropy loss:
\begin{equation}
\mathcal{L}_{\mathrm{pt}}
=
\sum_i
\sum_{m\in B_i}
w_{i,m}
\mathrm{BCE}
\left(
y_{i,m},
p_{\mathrm{pt}}(m\mid x_i)
\right),
\end{equation}
where $w_{i,m}$ alleviates medication-frequency imbalance.

To improve candidate ranking, we further construct positive-negative medication pairs and apply a pairwise ranking loss:
\begin{equation}
\begin{aligned}
\mathcal{L}_{\mathrm{pw}}
=
&\sum_i
\sum_{(m^+,m^-)\in\Pi_i}
\mathrm{softplus}
\Big(
-\big[
r_{\mathrm{pw}}(m^+\mid x_i)
\\
&\qquad\qquad\qquad
-
r_{\mathrm{pw}}(m^-\mid x_i)
\big]
\Big),
\end{aligned}
\end{equation}
where $\Pi_i=\{(m^+,m^-)\mid m^+\in M_i\cap B_i,\ m^-\in B_i\setminus M_i\}$.

Since HEM-Med predicts the medication set size rather than using a Top-$K$ threshold, the set-size predictor is trained by:
\begin{equation}
\mathcal{L}_{\mathrm{size}}
=
-
\sum_i
\log p_{\mathrm{size}}(|M_i|\mid x_i).
\end{equation}

Pointwise, pairwise, therapeutic-class, and medication-count outputs are combined only at inference for calibrated reranking and DDI-aware set decoding. DDI knowledge serves only candidate-level safety penalties and final decoding, not additional supervision.

\section{Experiment}
\subsection{Dataset}
\label{sec:dataset}

We evaluate HEM-Med on the eICU Collaborative Research Database, a real-world multi-center ICU database with over 200,000 de-identified ICU admissions in the United States~\cite{pollard2018eicu}. 
In the processed cohort, the ground-truth medication sets have a DDI Rate of approximately 11.66\%.
Dataset statistics are reported in Table~\ref{tab:eicu_statistics}.

\begin{table}[!h]
\centering
\caption{Statistics of the processed eICU dataset.}
 \label{tab:eicu_statistics}
\begin{tabular}{l r}
\hline
\textbf{Item} & \textbf{Number} \\
\hline
\# of records / hospitals & 90,545 / 149 \\
Avg. / max / min \# of records per hospital & 607.68 / 3,315 / 1 \\
Diag. / proc. / med. vocabulary size & 2,958 / 1,852 / 161 \\
Lab vocabulary size & 12 \\
Avg. / max / min \# of diagnoses per record & 4.04 / 91 / 1 \\
Avg. / max / min \# of procedures per record & 6.60 / 102 / 1 \\
Avg. / max / min \# of medications per record & 14.36 / 63 / 0 \\
Avg. / max / min \# of lab features per record & 3.76 / 12 / 0 \\
\# of records in train / validation / test & 72,436 / 9,054 / 9,055 \\
\hline
\end{tabular}
\vspace{-3mm}
\end{table}

\subsection{Evaluation Metrics}
\label{sec:metrics}

Following prior studies~\cite{shang2019gamenet,yang2021safedrug,yang2023molerec}, we assess accuracy using Jaccard and F1 for set overlap and PRAUC for prediction ranking, and safety using DDI Rate, the proportion of potentially interacting predicted medication pairs. 
To report  the overall performance, we introduce SafeScore:
\begin{equation}
\mathrm{SafeScore}
=
\frac{\mathrm{PRAUC}+\mathrm{F1}+\mathrm{Jaccard}}{3}
\cdot
(1-\mathrm{DDI}).
\end{equation}
Higher SafeScore indicates a better accuracy-safety trade-off.

\subsection{Implementation Details}

We split the dataset into training, validation, and test sets
with a ratio of 8:1:1. To avoid information leakage, all memory
statistics and prior distributions are constructed only from the
training set. For $\mathcal{K}^{(L_1)}$, the minimum anchor
support threshold $\tau_a$, anchor--medication support threshold
$\tau_{am}$, conditional frequency threshold $\tau_p$,
anchor--medication hospital-coverage threshold $\tau_h$, and
anchor hospital-coverage threshold $\tau_{ha}$ are set to
30, 10, 0.02, 4, and 5, respectively. For $\mathcal{K}^{(L_2)}$,
the local anchor support threshold $\tau^{(L_2)}_a$, local
anchor--medication support threshold $\tau^{(L_2)}_{am}$,
local conditional frequency threshold $\tau^{(L_2)}_p$, and
residual magnitude threshold $\tau^{(L_2)}_{\Delta}$ are set
to 8, 4, 0.10, and 0.12, respectively. These thresholds are
used as reliability filters for memory construction rather than
primary model-selection hyperparameters.

We set candidate depth $B=60$ and weak-experience decay factors $\gamma_{\mathrm{weak}}^{(L_1)}=\gamma_{\mathrm{weak}}^{(L_2)}=0.42$, and analyze their effects on candidate recall and noise. Other parameters are validation-selected and fixed during testing. All baselines share the same data split, medication vocabulary, DDI file~\cite{tatonetti2012data,wishart2018drugbank}, and random seed. Experiments run on a single NVIDIA GeForce RTX 3090 GPU with 24 GB memory.

\subsection{Result Analysis}
\label{sec:result_analysis}

\begin{table*}[!t]
\centering
\caption{Main experimental results on the eICU dataset.}
\label{tab:main_results}
\resizebox{0.9\textwidth}{!}{
\begin{tabular}{lccccc}
\toprule
\textbf{Model} & \textbf{PRAUC} $\uparrow$ & \textbf{F1} $\uparrow$ & \textbf{Jaccard} $\uparrow$ & \textbf{DDI Rate} $\downarrow$ & \textbf{SafeScore} $\uparrow$\\
\midrule
EMCDR      
& $0.3421 \pm 0.0043$ 
& $0.3716 \pm 0.0086$ 
& $0.2647 \pm 0.0095$ 
& $0.1321 \pm 0.0062$ 
& $0.2831 \pm 0.0044$ \\

DDTCDR     
& $0.3598 \pm 0.0136$ 
& $0.3763 \pm 0.0019$ 
& $0.2665 \pm 0.0102$ 
& $0.1231 \pm 0.0049$ 
& $0.2931 \pm 0.0053$ \\

TEMPT      
& $0.3430 \pm 0.0065$ 
& $0.3335 \pm 0.0054$ 
& $0.2284 \pm 0.0075$ 
& $0.1204 \pm 0.0038$ 
& $0.2653 \pm 0.0035$ \\

Reason4Rec 
& $0.3697 \pm 0.0083$ 
& $0.3893 \pm 0.0022$ 
& $0.2783 \pm 0.0036$ 
& $\mathbf{0.0814 \pm 0.0064}$ 
& $0.3176 \pm 0.0036$ \\

EXP3RT     
& $0.3732 \pm 0.0148$ 
& $0.3944 \pm 0.0026$ 
& $0.2819 \pm 0.0028$ 
& $0.0927 \pm 0.0014$ 
& $0.3174 \pm 0.0046$ \\

SafeDrug   
& $\underline{0.3926 \pm 0.0038}$ 
& $\underline{0.4082 \pm 0.0061}$ 
& $\underline{0.2939 \pm 0.0049}$ 
& $0.0862 \pm 0.0064$ 
& $\underline{0.3334 \pm 0.0035}$ \\

VITA       
& $0.3578 \pm 0.0037$ 
& $0.3782 \pm 0.0025$ 
& $0.2688 \pm 0.0120$ 
& $0.1107 \pm 0.0071$ 
& $0.2979 \pm 0.0045$ \\

RES-MR     
& $0.3697 \pm 0.0046$ 
& $0.3893 \pm 0.0019$ 
& $0.2782 \pm 0.0076$ 
& $0.1013 \pm 0.0021$ 
& $0.3107 \pm 0.0028$ \\

\textbf{HEM-Med} 
& $\mathbf{0.4161 \pm 0.0086}$ 
& $\mathbf{0.4532 \pm 0.0062}$ 
& $\mathbf{0.3246 \pm 0.0074}$ 
& $\underline{0.0831 \pm 0.0035}$ 
& $\mathbf{0.3649 \pm 0.0042}$ \\
\bottomrule
\end{tabular}
}
\begin{minipage}{0.9\textwidth}
\footnotesize
\ The best results are marked in bold and the second-best results are underlined.
\end{minipage}
\end{table*}

\begin{table*}[!ht]
\centering
\caption{Ablation study results on the eICU dataset.}
\label{tab:ablation_results}
\resizebox{0.9\textwidth}{!}{
\begin{tabular}{lccccc}
\toprule
\textbf{Variant} & \textbf{PRAUC} $\uparrow$ & \textbf{F1} $\uparrow$ & \textbf{Jaccard} $\uparrow$ & \textbf{DDI Rate} $\downarrow$ & \textbf{SafeScore} $\uparrow$\\
\midrule
HEM-Med w/o MD 
& $0.3723 \pm 0.0145$ 
& $\underline{0.4221 \pm 0.0019}$ 
& $\underline{0.2938 \pm 0.0092}$ 
& $0.0993 \pm 0.0187$ 
& $\underline{0.3267 \pm 0.0085}$ \\

HEM-Med w/o DDI-Dec 
& $0.3631 \pm 0.0136$ 
& $0.4175 \pm 0.0021$ 
& $0.2840 \pm 0.0102$ 
& $0.1254 \pm 0.0074$ 
& $0.3104 \pm 0.0056$ \\

HEM-Med w/o RERANKER 
& $\underline{0.4025 \pm 0.0049}$ 
& $0.3235 \pm 0.0078$ 
& $0.2111 \pm 0.0053$ 
& $0.0953 \pm 0.0049$ 
& $0.2826 \pm 0.0036$ \\

HEM-Med w/o WAM 
& $0.3677 \pm 0.0047$ 
& $0.4117 \pm 0.0091$ 
& $0.2824 \pm 0.0062$ 
& $\underline{0.0886 \pm 0.0037}$ 
& $0.3226 \pm 0.0039$ \\

\textbf{HEM-Med} 
& $\mathbf{0.4161 \pm 0.0086}$ 
& $\mathbf{0.4532 \pm 0.0062}$ 
& $\mathbf{0.3246 \pm 0.0074}$ 
& $\mathbf{0.0831 \pm 0.0035}$ 
& $\mathbf{0.3649 \pm 0.0042}$ \\
\bottomrule
\end{tabular}
}
\begin{minipage}{0.9\textwidth}
\footnotesize
\end{minipage}
\vspace{-3mm}
\end{table*}

Table~\ref{tab:main_results} reports results on the preprocessed eICU dataset. Baselines include transfer learning methods EMCDR and DDTCDR~\cite{man2017cross,li2020ddtcdr}, general LLM recommenders Reason4Rec and EXP3RT~\cite{fang2025reason4rec,kim2025driven}, and MR methods SafeDrug, VITA, TEMPT, and RES-MR~\cite{yang2021safedrug,kim2024vita,liu2025contrastive,wang2026resmr}. TEMPT targets multi-center MR. All methods undergo 10 bootstrap evaluations with a sampling ratio of 0.8; metric means and standard deviations are reported.

HEM-Med surpasses transfer and general LLM baselines in accuracy and achieves the highest SafeScore. It improves the accuracy--safety trade-off over SafeDrug through disentangled memory, calibrated reranking, and DDI-aware decoding. Reason4Rec yields the lowest DDI Rate but lower accuracy, highlighting the limitations of DDI reduction alone.

\subsection{Ablation Study}
\label{sec:ablation}
Table~\ref{tab:ablation_results} reports four ablations: w/o MD, w/o DDI-Dec, w/o RERANKER, and w/o WAM, removing memory decoupling, DDI-aware decoding, experience reranking, and weak auxiliary memory, respectively. Each removal degrades performance, supporting the contributions of memory disentanglement, safety-aware decoding, candidate calibration, and weak experience. Full HEM-Med achieves the best overall performance.

\subsection{Parameter Analysis}

\begin{figure}[htbp]
\centering

\begin{subfigure}{0.48\columnwidth}
    \centering
    \includegraphics[width=\linewidth]{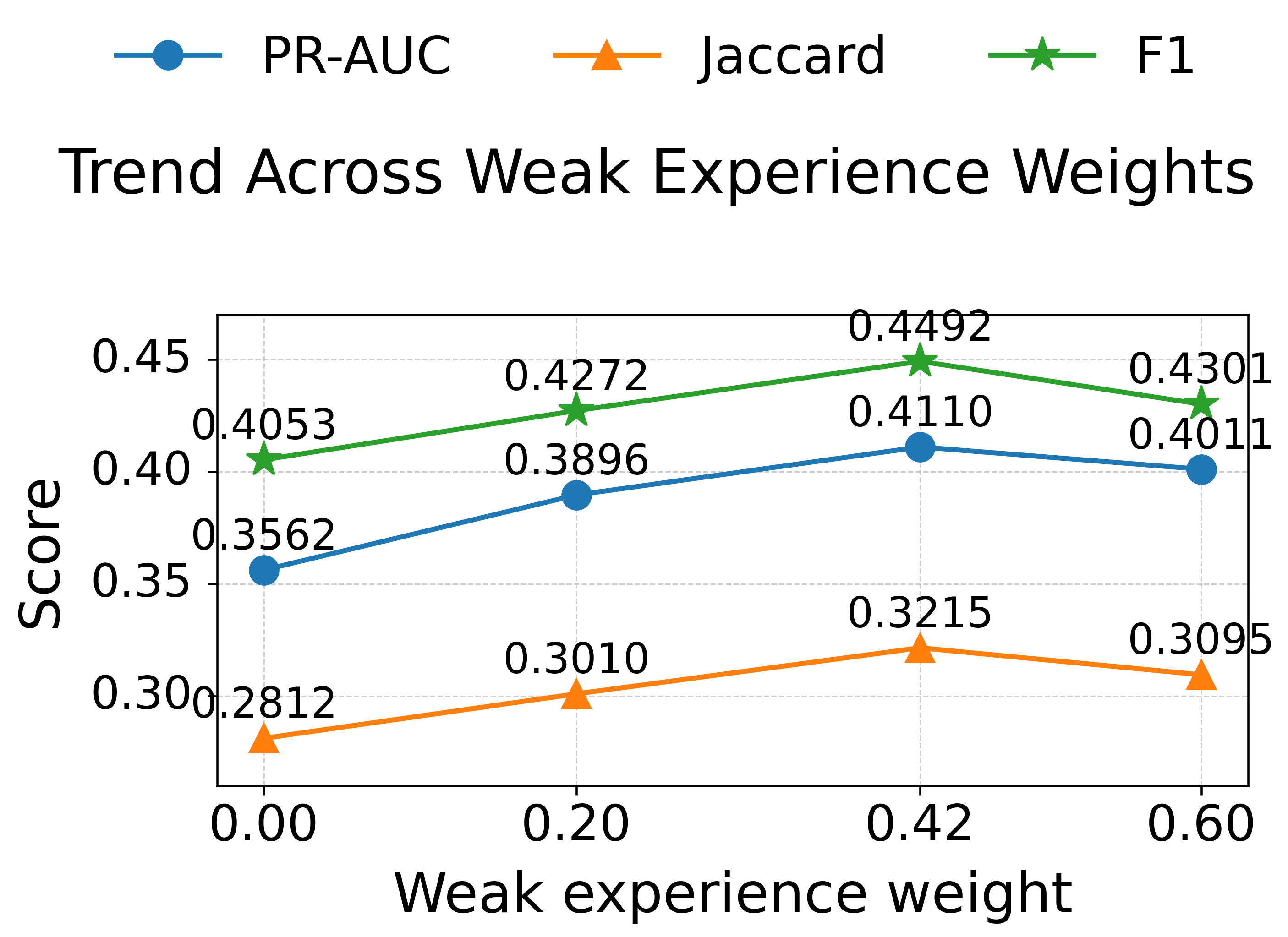}
    \caption{Weak weight}
    \label{fig:weak_weight}
\end{subfigure}
\hfill
\begin{subfigure}{0.48\columnwidth}
    \centering
    \includegraphics[width=\linewidth]{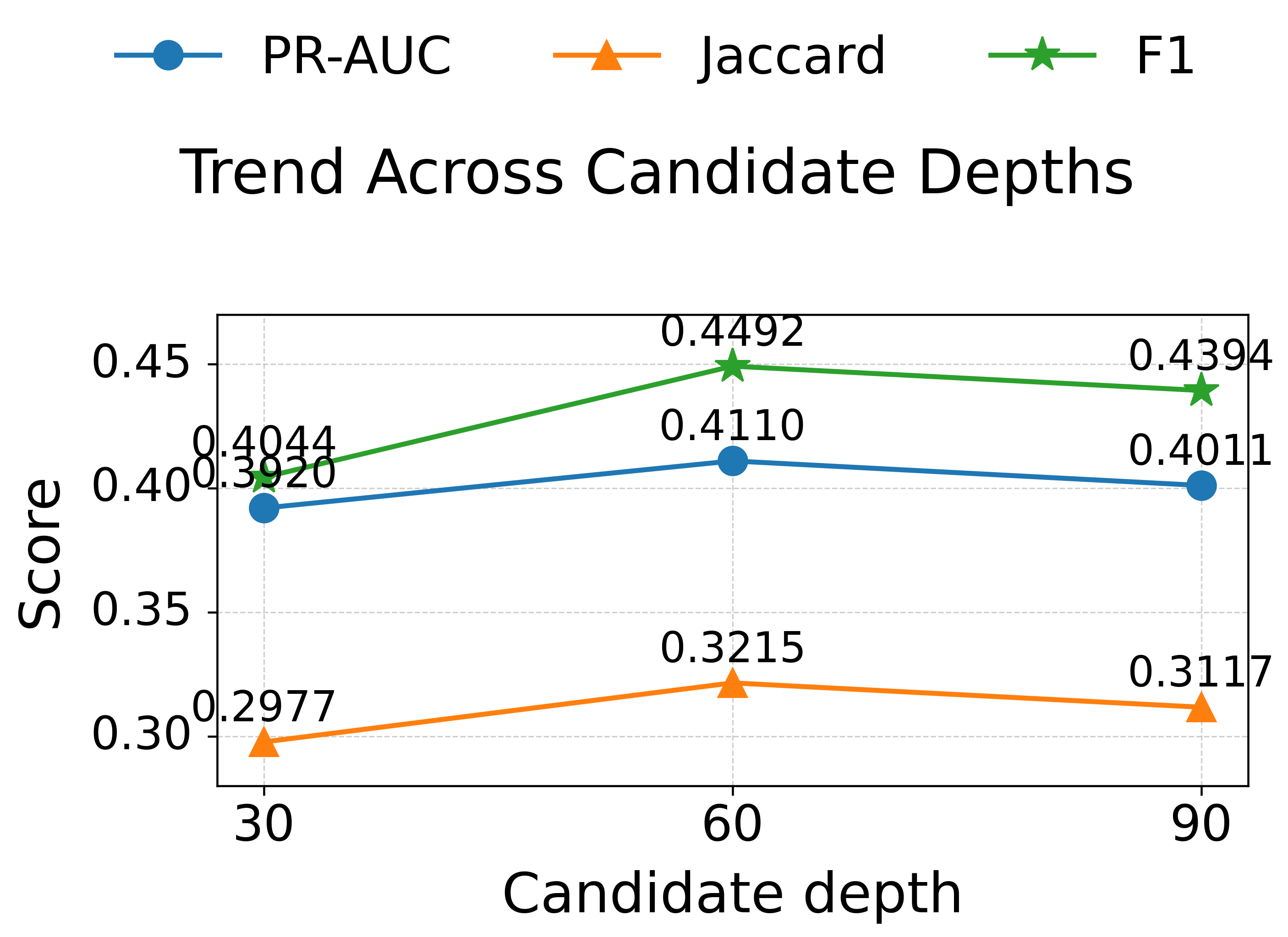}
    \caption{Candidate depth}
    \label{fig:candidate_depth}
\end{subfigure}

\caption{Parameter analysis of weak experience weight and candidate depth.}
\label{fig:parameter_analysis}
\vspace{-1mm}
\end{figure}

We analyze the influence of two key hyperparameters: the weak auxiliary experience weight and the candidate depth. 
The results are shown in Fig.~\ref{fig:parameter_analysis}.

\textbf{Impact of weak auxiliary experience weight.}
With the candidate depth fixed to 60, introducing weak auxiliary experience improves recommendation performance compared with removing weak memory. The model achieves the best performance when the weak experience weight is set to 0.42. Further increasing the weight leads to performance degradation, indicating that weak experience is useful as supplementary information but may introduce noise when overemphasized.

\textbf{Impact of candidate depth.}
With the weak experience weight fixed to 0.42, the model performs best when the candidate depth is set to 60. 
A smaller candidate depth may miss relevant medications before reranking, while a larger depth introduces more noisy candidates and false positives. 
Therefore, we set the weak auxiliary experience weight to 0.42 and the candidate depth to 60 in the final experiments.



\subsection{Case Study}

Table~\ref{tab:case_two_hospitals_single} presents two ICU cases from different hospitals. Both cases involve alcohol or toxicology-related conditions and electrolyte management, yet their medication requirements differ. Although the baseline methods identify some common electrolyte medications, they frequently produce false-positive recommendations and miss key medications, such as \textit{Furosemide} in Case A and \textit{Hydralazine} in Case B, indicating limited adaptation to hospital- and patient-specific needs. In contrast, HEM-Med accurately matches the medication sets in both cases, preserving shared electrolyte medications while identifying case-specific requirements. This demonstrates its ability to integrate cross-hospital common knowledge, hospital-specific experience, and safety-aware decoding.

\begin{table}[t]
\centering
\caption{Case study comparison between two clinically similar ICU stays from different hospitals.}
\label{tab:case_two_hospitals_single}
\scriptsize
\setlength{\tabcolsep}{2pt}
\renewcommand{\arraystretch}{1.12}
\begin{tabularx}{\columnwidth}{@{}p{0.19\columnwidth}YY@{}}
\toprule
\textbf{Method} 
& \textbf{Case A: H452} 
& \textbf{Case B: H444} \\
\midrule

Ground Truth
& \ \  Dex; Fur; NaCl; KCl; MgSO$_4$
& \  \ Enox; Hyd; Lor; NaCl; KCl \\

\midrule
DDTCDR
& \predcell{3/5}{
\casehit{NaCl}; \casehit{MgSO$_4$}; \casefp{CaGlu}; \casehit{KCl}; \casefp{APAP}
}{
\casemiss{Dex; Fur}
}
& \predcell{3/5}{
\casehit{KCl}; \casefp{Ond}; \casefp{APAP}; \casehit{NaCl}; \casehit{Lor}
}{
\casemiss{Enox; Hyd}
} \\

SafeDrug
& \predcell{4/5}{
\casehit{NaCl}; \casefp{CaGlu}; \casehit{MgSO$_4$}; \casehit{KCl}; \casehit{Dex}
}{
\casemiss{Fur}
}
& \predcell{4/5}{
\casehit{KCl}; \casehit{Lor}; \casefp{Eso}; \casehit{NaCl}; \casehit{Enox}
}{
\casemiss{Hyd}
} \\

TEMPT
& \predcell{3/5}{
\casehit{NaCl}; \casefp{Lor}; \casehit{Dex}; \casehit{KCl}; \casefp{Glu}
}{
\casemiss{Fur; MgSO$_4$}
}
& \predcell{4/5}{
\casehit{Enox}; \casehit{Lor}; \casehit{NaCl}; \casehit{KCl}; \casefp{Eso}
}{
\casemiss{Hyd}
} \\

RES-MR
& \predcell{4/5}{
\casehit{NaCl}; \casehit{KCl}; \casehit{MgSO$_4$}; \casefp{CaGlu}; \casehit{Dex}
}{
\casemiss{Fur}
}
& \predcell{4/5}{
\casehit{KCl}; \casehit{Lor}; \casehit{NaCl}; \casefp{Eso}; \casehit{Enox}
}{
\casemiss{Hyd}
} \\

\textbf{HEM-Med}
& \predcell{5/5}{
\textbf{\casehit{NaCl}; \casehit{KCl}; \casehit{MgSO$_4$}; \casehit{Dex}; \casehit{Fur}}
}{
None
}
& \predcell{5/5}{
\textbf{\casehit{Lor}; \casehit{Enox}; \casehit{KCl}; \casehit{Hyd}; \casehit{NaCl}}
}{
None
} \\

\bottomrule
\multicolumn{3}{@{}p{\columnwidth}@{}}{
\scriptsize
\textit{Color legend:}
\casehit{Blue} = correctly predicted medication;
\casefp{Red} = false-positive medication;
\casemiss{Orange} = missed ground-truth medication.
}
\\[-1pt]
\multicolumn{3}{@{}p{\columnwidth}@{}}{
\scriptsize
\textit{Abbreviations:}
Dex = Dextrose; Fur = Furosemide; NaCl = Sodium chloride; KCl = Potassium chloride;
MgSO$_4$ = Magnesium sulfate; CaGlu = Calcium gluconate; APAP = Acetaminophen;
Ond = Ondansetron; Eso = Esomeprazole; Glu = Glucagon; Enox = Enoxaparin;
Hyd = Hydralazine; Lor = Lorazepam.
}
\end{tabularx}
\vspace{-3mm}
\end{table}

\section{Conclusion}

In this paper, we propose HEM-Med, a hierarchical agentic memory framework for safe multi-center medication recommendation. HEM-Med explicitly separates cross-hospital consensus experience from hospital-specific residual patterns and integrates DDI-aware decoding for safer medication set generation.
Experiments on a real-world multi-center eICU dataset show that HEM-Med improves recommendation  accuracy while maintaining medication safety. Ablation studies validate the effectiveness of its key components. Future work will explore richer clinical constraints, fine-grained medication safety knowledge, and prospective validation.

\section*{Acknowledgment}This work is supported by the National Natural Science Foundation of China (No. 62376130), Program of New Twenty Policies for Universities of Jinan (No. 202333008), the Pilot Project for Integrated Innovation of Science, Education, and Industry of Qilu University of Technology (Shandong Academy of Sciences) (No. 2025ZDZX01).

\bibliographystyle{IEEEtran}
\bibliography{references}

\end{document}